\documentclass{article}

\usepackage[utf8]{inputenc} 
\usepackage[T1]{fontenc}    
\usepackage{url}            
\usepackage{booktabs}       
\usepackage{amsfonts}       
\usepackage{nicefrac}       
\usepackage{amsmath}
\usepackage{graphicx}
\usepackage{subcaption}
\usepackage{amssymb}

\newcommand{\R}{\mathbb{R}} 

\newcommand{\q}{q} 
\newcommand{\A}{A} 
\newcommand{\ac}{a} 
\newcommand{\B}{B} 
\newcommand{\bb}{b} 

\newcommand{\sens}[1]{\tilde{#1}}
\newcommand{\adjoint}[1]{\overline{#1}}

\newcommand{\conv}{*}
\newcommand{\ccorr}{\star}

\newcommand{\tvec}[1]{\mathbf{#1}}
\newcommand{\nsamp}{T}
\newcommand{\pdiff}[2]{\frac{\partial #1}{\partial #2}}

\newcommand{\loss}{\mathcal{L}}

\newcommand{\Name}{\emph{dynoNet}}

\usepackage{listings}
\DeclareFixedFont{\ttb}{T1}{txtt}{bx}{n}{12} 
\DeclareFixedFont{\ttm}{T1}{txtt}{m}{n}{12}  

\usepackage{color}
\definecolor{deepblue}{rgb}{0,0,0.5}
\definecolor{deepred}{rgb}{0.6,0,0}
\definecolor{deepgreen}{rgb}{0,0.5,0}
\newcommand\pythonstyle{\lstset{
language=Python,
basicstyle=\ttb\tiny,
otherkeywords={self},             
keywordstyle=\tiny\color{deepblue},
emph={MyClass,__init__},          
emphstyle=\ttb\color{deepred},    
stringstyle=\color{deepgreen},
frame=tb,                         
showstringspaces=false,            %
numberstyle=\footnotesize,
}}

\lstnewenvironment{python}[1][]
{
\pythonstyle
\lstset{#1}
}
{}

\newcommand\pythoninline[1]{{\pythonstyle\lstinline!#1!}}

\title{\Name: A neural network architecture for learning dynamical systems}

\author{%
  Marco~Forgione, Dario Piga\\
  IDSIA Dalle Molle Institute for Artificial Intelligence\\
  USI-SUPSI, Lugano, Switzerland\\
  \texttt{\{marco.forgione, dario.piga\}@supsi.ch} \\
}

\begin{document}

\maketitle
\abstract{This paper introduces a  network architecture, called \Name, utilizing linear dynamical operators as elementary building blocks.  Owing to the dynamical nature of these blocks,  \Name \  networks are tailored for sequence modeling and system identification purposes. The back-propagation behavior of the linear dynamical operator with respect to both its parameters and its input sequence is defined. This enables end-to-end training of structured networks containing linear dynamical operators and other differentiable units,  exploiting existing deep learning software. Examples show the effectiveness of the proposed approach on well-known system identification benchmarks.
}

\vskip 1em 
\noindent\rule{\textwidth}{1pt}
 To cite this work, please use the following bibtex entry:
\begin{verbatim}
@article{forgione2021a,
  title={\textit{dyno{N}et}: A neural network architecture for learning dynamical systems},
  author={Forgione, M. and Piga, D.},
  journal={International Journal of Adaptive Control and Signal Processing},
  volume={35},
  number={4},
  pages={612--626},
  year={2021},
  publisher={Wiley}
}
\end{verbatim}
\vskip 1em
Using the plain bibtex style, the bibliographic entry should look like:\\ \\
\textsf{M. Forgione and D. Piga. \textit{dynoNet}: A neural network architecture for learning
dynamical systems.} \textit{International Journal of Adaptive Control and Signal Processing}, 35(4):612--626, 2021.

\noindent\rule{\textwidth}{1pt}

\section{Introduction}
\subsection{Contribution}
This paper introduces \Name, a neural network architecture tailored for sequence modeling and dynamical system learning (a.k.a. \emph{system identification}). The network is designed to process time series of arbitrary length and contains causal \emph{linear time-invariant} (LTI) dynamical operators as building blocks. 
These LTI layers are parametrized in terms of  rational transfer functions, and thus apply \emph{infinite impulse response} (IIR) filtering to their input sequence. In the \Name\ architecture, the LTI layers are combined with \emph{static} (i.e., memoryless) non-linearities which can be  either elementary \emph{activation functions} applied {channel-wise}; fully connected feed-forward neural networks; or other differentiable operators (e.g, polynomials). 
Both the LTI and the static layers defining a  \Name\ are in general \emph{multi-input-multi-output} (MIMO) and can be interconnected in an arbitrary fashion. 

Overall, the \Name \ architecture can represent rich classes of non-linear, causal dynamical relations. Moreover,  \Name \ networks can be trained \emph{end-to-end} by plain \emph{back-propagation} using standard \emph{deep learning} (DL) software. Technically, this is achieved by introducing the LTI dynamical layer as a differentiable operator, endowed with a well-defined forward and backward behavior and thus compatible with reverse-mode automatic differentiation \cite{baydin2017automatic}.  
Special care is taken to devise closed-form expressions for the forward and backward operations that are convenient from a computational perspective.

A software implementation of the linear dynamical operator based on the \emph{PyTorch} DL framework \cite{paszke:2017automatic} 
has been developed and is available in the GitHub repository \url{https://github.com/forgi86/dynonet.git}.

\subsection{Related works}
\label{sec:related}

To the best of our knowledge, LTI blocks with an IIR have never been considered as differentiable operators for back-propagation-based training to date.
Among the layers routinely applied in DL, 1-D convolution \cite{wang2017time} is the closest match. In particular, the 1D causal convolution layer\cite{bai2018empirical,andersson:2019deep} corresponds to the filtering of an input sequence through a causal \emph{finite impulse response} (FIR) dynamical system.  
The \Name\ architecture may be seen as a generalization of the causal 1D \emph{convolutional neural network} (CNN) enabling IIR filtering, owing to the description of the dynamical layers as rational transfer functions. 
This representation allows modeling long-term (actually infinite) time dependencies with a smaller number of parameters with respect to 1D convolutional networks. 
Furthermore, filtering through rational transfer function can be implemented by means of recurrent linear difference equations. While this operation is not as highly parallelizable as FIR filtering, the total number of computations required is generally lower.

The \Name \  architecture has also analogies with \emph{recurrent neural network} (RNN) \cite{greff2016lstm} architectures. As in RNNs, a dynamic dependency is built exploiting recurrence equations. However, in  an RNN the basic computational unit is a neural cell (e.g., Elman, LSTM, GRU) that processes a single time step. The network's computational graph is then built by repeating the same cell for all the steps of the  timeseries. Processing long timeseries through an RNN is often computational expensive and presents limited opportunities for parallelization, due to the sequential structure of the computational graph.
Conversely, in \Name  \ a lightweight (linear) recurrence equation is ``baked into'' the elementary LTI blocks, that naturally operate on time series in a vectorized fashion.
While internally these layers require certain sequential operations (details are given in Section \ref{sec:layer_DL}), the overall computational burden is sensibly lower than the one of typical RNNs.
Moreover, the computations performed by the static layers of a \Name\ are highly parallelizable, as they are independent for each time step. Therefore, more complex transformations may be included in the static layers. 

Thus, compared to 1D convolutional and recurrent neural architectures, \Name \ is characterized by an intermediate level of computational complexity and flexibility.

\begin{figure}[h]
	\centering
	\begin{subfigure}{.4\textwidth}
		\centering
		\includegraphics[width=.6\linewidth]{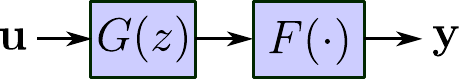}
		\subcaption{Wiener}
	\end{subfigure}%
	\begin{subfigure}{.4\textwidth}
		\centering
		\includegraphics[width=.6\linewidth]{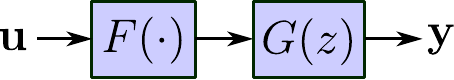}
		\subcaption{Hammerstein}
	\end{subfigure}
	\vskip .5em
	\begin{subfigure}{.4\textwidth}
		\centering
		\includegraphics[width=.8\linewidth]{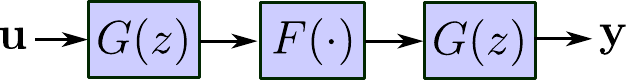}
		\subcaption{Wiener-Hammerstein}
	\end{subfigure}%
	\begin{subfigure}{.4\textwidth}
		\centering
		\includegraphics[width=.8\linewidth]{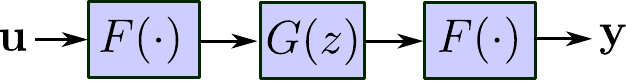}
		\subcaption{Hammerstein-Wiener}
	\end{subfigure}
	\caption{Classic block-oriented architectures. All blocks are SISO.}
	\label{fig:classic_blockoriented}
\end{figure}

In the {system identification literature}, particular cases of the \Name \ architecture have been widely studied in the last decades within the so-called \emph{block-oriented} modeling framework \cite{giri2010block}.
In most of the contributions, shallow architectures based on \emph{single-input single-output} (SISO) blocks are considered.
For instance:  the Wiener model is defined as the series connection of an LTI dynamical model $G$ followed by a static non-linearity $F$;  the Hammerstein model is based on the reverse connection, with a static non-linearity $F$ followed by an LTI block $G$; the Wiener-Hammerstein (WH) model combines two SISO LTI blocks interleaved by a static SISO non-linearity in a sequential structure $G$-$F$-$G$; and the Hammerstein-Wiener (HW) has  structure $F$-$G$-$F$.  See Figure \ref{fig:classic_blockoriented} for a visual representation of the aforementioned structures. 

\begin{figure}
	\centering
	\begin{subfigure}{.45\textwidth}
		\vskip -0.6em
		\centering
		\includegraphics[width=.99\linewidth]{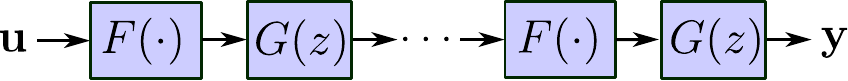}
		\label{fig:sub1}
	\end{subfigure}%
	\begin{subfigure}{.2\textwidth}
	\end{subfigure}
	\begin{subfigure}{.45\textwidth}
		\centering
		\vskip 2em
		\includegraphics[width=.9\linewidth]{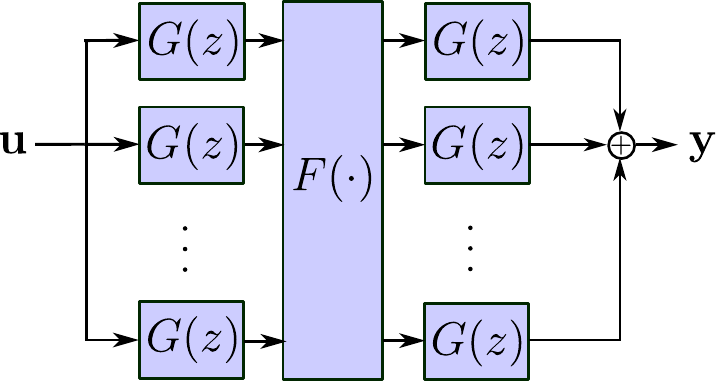}
		\label{fig:sub2}
	\end{subfigure}
	\caption{Generalized Hammerstein-Wiener (left) and Parallel Wiener-Hammerstein (right) model structures.}
	\label{fig:medium_blockoriented}
\end{figure}

One   exception is the deep architecture consisting in the repeated sequential connection of SISO blocks $F$-$G$-$F$-$G$ $\dots$ described in \cite{wills:2012generalised} and called  generalized Hammerstein-Wiener (Figure \ref{fig:medium_blockoriented}, left panel). Furthermore, the parallel Wiener-Hammerstein model \cite{schoukens2015parametric} extends the classic WH model beyond the strictly SISO case. 
Indeed, the parallel WH model has the same $G$-$F$-$G$ as the basic WH mentioned above. However, the first  linear block is single-input-multi-output; the static non-linearity $F$ is multi-input-multi-output; and the second linear block is multi-input-single-output (Figure \ref{fig:medium_blockoriented}, right panel). Overall, the parallel WH model describes an input/output SISO dynamical system, but it leverages on an inner MIMO structure to provide additional flexibility.
From a DL perspective, the parallel WH extends the representation capabilities of the plain WH network by including several ``neurons'' in a single hidden layer, while the generalized HW model aims to the same result by stacking several layers, each one consisting of a single ``neuron''.

The \Name\ architecture encompasses all the previous block-oriented models as special cases. Other structures containing, e.g., multiple MIMO blocks and \emph{skip connection} can be described within the \Name  \ modeling framework.
More importantly,  the existing training methods for block-oriented models are custom-made for each specific architecture, requiring for instance analytic expressions of the Jacobian of the loss with respect to the training parameters. Conversely, the derivation of the differentiable dynamical layer allows us to train arbitrary \Name \  architectures using the plain back-propagation algorithm. 

\subsection{Representational power of \emph{dynoNet} architectures}
The \Name\ framework for dynamical systems is at least as expressive as the block-oriented models in Figures~\ref{fig:classic_blockoriented} and~\ref{fig:medium_blockoriented}, and causal CNN architectures, being a generalization of the two approaches.

Formal results on the representational power of block-oriented models are given in \cite{palm:1979representation, boyd:1985fading}. It is proven that certain block-oriented models are general approximators of non-linear \emph{fading memory} operators. In particular, the property holds for ``Parallel Wiener'' architectures consisting in a  single-input-multi-output LTI block followed by a multi-input-single-output static non-linearity. 

Note that fading memory operators may describe a wide range of non-linear phenomena, but they exclude the cases of time-varying, chaotic, finite-escape-time, unstable, and multiple-equilibria systems. While \Name\ can also describe certain unstable systems
(rational transfer functions may have unstable dynamics) it is hard to further characterize the \Name\ model 
structure from a system theoretic perspective.

From a practical perspective, the effectiveness of deep CNN architectures for different system identification and 
time series modeling tasks has been demonstrated in several contributions \cite{ wang2017time,bai2018empirical,andersson:2019deep}. 
Even though mathematically it is not clear whether increasing the number of hidden layers extends the class of dynamics that can be represented by CNNs, experimentally it has been observed that deeper networks are able to learn more complex dependencies than shallower ones, for a given number of training parameters and for a given computational effort.

While the \Name\ framework does not increase the theoretical representation power of CNNs dramatically (one could argue that any stable \Name\ architecture may be approximated arbitrarily well with a causal CNN, by including convolutional units with a sufficiently long memory), it allows representing long-term dependencies with a smaller number of parameters. This 
may lead to substantial advantages in terms of generalization properties, memory requirements, training and inference time of the models.

\subsection{Notation}
The following notation will be used throughout the paper. 
The entries of an $n$-length vector are specified by subscript integer indices running from $0$ to $n\!-\!1$, unless stated otherwise. 
The bold-face notation is reserved for real-valued $\nsamp$-length vectors, generally representing time series with $T$ samples. 
For instance, $\tvec{u} \in \R^{\nsamp}$ is a $T$-length vector with entries $\tvec{u}_0, \tvec{u}_1, \dots, \tvec{u}_{\nsamp -1}$.

\paragraph{Time reversal}
The time reversal of a $\nsamp$-length vector $\tvec{u} \in \R^{\nsamp}$ is denoted as $\textrm{flip}(\tvec{u})$ and defined as
\begin{equation}
\left(\textrm{flip}(\tvec{u})\right)_t = \tvec{u}_{\nsamp-t-1},\qquad t=0,1,\dots,\nsamp-1
\end{equation}
\paragraph{Convolution}
The convolution between vectors ${x}\in \R^{n_x}$ and ${y} \in \R^{n_y}$ is defined as 
\begin{equation}
(x \conv y)_i = \sum_{j=\max{(0, i+1-n_y)}}^{\min(i, n_x-1)} x_j y_{i-j},\qquad i=0,1,\dots,n_x+n_y-1.
\end{equation}
\paragraph{Cross-correlation}
The cross-correlation between vectors ${x}\in \R^{n_x}$ and 
${y} \in \R^{n_y}$ is  defined as 
\begin{equation}
(x \ccorr y)_i = \sum_{j=\max(i,0)}^{\min(n_x+i-1, n_y-1)} x_{j-i} y_{j},\qquad i=-n_x+1,\dots,n_y-1.
\end{equation}

\section{Linear dynamical operator} 
\label{sec:LTI}
The input-output relation of an individual (SISO) dynamical layer in the \Name \ architecture is described by the dynamical rational operator $G(q)$ 
as follows:
\begin{subequations}
	\label{eq:OE_predictor_a}
	\begin{equation} \label{eqn:filter}
	y(t) = G(\q)u(t) = \frac{\B(\q)}{\A(\q)} u(t),
	\end{equation}
	where $A(q)$ and $B(q)$ are polynomials in the \emph{time delay operator}  $\q^{-1}$ ($q^{-1}u(t)=u(t-1)$), i.e., 
	\begin{align}
	\A(q) &= 1 + \ac_1 \q^{-1} + \dots + \ac_{n_\ac}q^{-n_\ac}, \\
	\B(q) &= \bb_0 + \bb_1 \q^{-1} + \dots + \bb_{n_{\bb}}q^{-n_{\bb}},
	\end{align}
	and  $u(t) \in \R$ and  $y(t) \in \R$ are the input and output sequence values at time index $t$.
\end{subequations}

The filtering operation through $G(\q)$ in \eqref{eqn:filter} is equivalent to the input/output equation:
\begin{equation} \label{eqn:filterA}
\A(\q)y(t) =  \B(\q)u(t).
\end{equation}
Based on the definitions of $\A(q)$ and $\B(q)$,~\eqref{eqn:filterA} is equivalent to the recurrence equation:
\begin{equation}
\label{eq:OE_predictor_b}
y(t) = \bb_0 u(t) + \bb_1 u(t-1) + \dots + \bb_{n_\bb}\!u(t-n_\bb)-\ac_1 y(t\!-\!1) \dots - \ac_{n_\ac} y(t\!-\!n_\ac).
\end{equation}

\paragraph{Parameters} The tunable parameters of $G(\q)$ are the coefficients of the polynomials $\A(q)$ and $\B(q)$. For convenience, these coefficients are collected in vectors  
$\ac = [\ac_1\; \ac_2\dots\;\ac_{n_\ac}] \in \R^{n_\ac}$ and $\bb = [\bb_0\; \bb_1\; \dots \;\bb_{n_\bb}] \in \R^{n_\bb + 1}$. 
Note that the first element of vector $\ac$ has index 1, while for all other vectors in this paper the starting index is 0.

\paragraph{Initial condition} In this paper, the operator $G(\q)$ is always initialized from rest, namely the values of $u(t)$ and $y(t)$ for $t < 0$ are all taken equal to zero. Then, given an input sequence $\{u(t),\; t \geq 0\}$, \eqref{eq:OE_predictor_b} provides an univocal expression for the output sequence $\{y(t),\; t \geq 0\}$.

\paragraph{Finite-length sequences} In practice, the operator $G(\q)$ in a \Name\ operates on finite-length sequences. Let us stack the input and output samples $u(t)$ and $y(t)$ in vectors $\tvec{u} \in \mathbb{R}^{\nsamp}$  and $\tvec{y} \in \mathbb{R}^{\nsamp}$, respectively. 
With a slight abuse of notation, the filtering operation in \eqref{eq:OE_predictor_a} applied to $\tvec{u}$ is denoted as
\begin{equation*}
\tvec{y} = G(\q)\tvec{u}.
\end{equation*}
The operation above is also equivalent to the convolution 
\begin{equation}
\label{eq:G_conv}
\tvec{y}_i = (\tvec{g} \conv \tvec{u})_i, \qquad i=0,1,\dots,\nsamp\!-\!1,
\end{equation}
where $\tvec{g} \in \R^{\nsamp}$ is a vector containing the first $\nsamp$ samples of the operator's \emph{impulse response}. The latter is defined as the output sequence generated by \eqref{eq:OE_predictor_b} for an input $u(\cdot)$ such that $u(0)=1$ and $u(t)=0, \forall t \geq 1$.

\paragraph{MIMO extension} 
In the MIMO case, the input $u(t) \in \R^{p}$ and output $y(t) \in \R^{m}$ at time $t$ are \emph{vectors} of size 
$p$ and $m$, respectively. The MIMO linear dynamical operator with $p$ input and $m$ output channels may be represented as a $m \times p$ MIMO transfer function matrix $G(\q)$ whose element $G_{kh}(\q)$ is a SISO rational transfer function such as \eqref{eq:OE_predictor_a}.
The components $y_k(t)$ of the output sequence $y(t)$ at time $t$ are defined as
\begin{equation}
y_k(t) = \sum_{h=0}^{p-1} G_{kh}(\q)u_h(t),\qquad k=0,1,\dots m-1. 
\end{equation}

The derivations for the dynamical layer are presented in the following in a SISO setting to avoid notation clutter. Extension to the MIMO case is  straightforward and  only requires repetition of the same operations for the different input/output channels. 
The computations for the different input/output channels are independent and therefore  may be performed in parallel. 

Note that the software implementation of the operator available in our on-line GitHub repository fully supports the MIMO case.

\section{Dynamical operator as a deep learning layer}
\label{sec:layer_DL}

In this section, the forward and backward operations required to integrate the linear dynamical operator in a DL framework are derived. The computational cost of these operations as measured by the number of multiplications to be executed is also reported.  Furthermore, the possibility of parallelizing these computations is analyzed.

In the rest of this paper, the linear dynamical operator interpreted as a differentiable layer for use in DL is also referred to as $G$-block. In our software implementation, the $G$-block is implemented in the \emph{PyTorch} DL framework
as a class extending \texttt{torch.autograd.Function}, based on the forward and backward operations derived in the following.

\subsection{Forward operations}
\begin{figure}[h]
	\begin{center}
		\includegraphics[width=140pt]{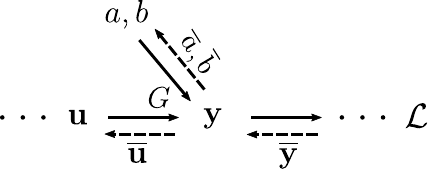}
	\end{center}
	\caption{Forward and backward operations of a $G$-block within a computational graph.}
	\label{fig:backprop_tf}
\end{figure}

The forward operations of a $G$-block embedded in a computational graph are represented by solid arrows in Figure \ref{fig:backprop_tf}.
In the {forward pass}, the block filters an input sequence $\tvec{u} \in \mathbb{R}^{\nsamp}$ through a dynamical system $G(\q)$ with structure \eqref{eq:OE_predictor_a} and parameters $\ac = [\ac_1\;\dots\;\ac_{n_\ac}]$ and $\bb = [\bb_0\; \bb_1\; \dots \;\bb_{n_\bb}]$. The block output is a vector $\tvec{y} \in \mathbb{R}^{\nsamp}$ containing the filtered sequence:
\begin{equation}
\label{eq:forward_op}
\tvec{y} = G.{\rm forward}(\tvec{u}, \bb, \ac) = G(\q) \tvec{u}.
\end{equation}
The input $\tvec{u}$ of the $G$-block  may be either the training input sequence or the result of previous operations in the computational graph, while
the output $\tvec{y}$ is an intermediate step towards the computation of a scalar output $\loss$. The exact operations leading to $\loss$ are not relevant in this discussion, and thus they are not further specified.

When the filtering operation \eqref{eq:forward_op} is implemented using \eqref{eq:OE_predictor_b}, the computational cost of the $G$-block forward pass corresponds to $\nsamp(n_\bb + n_\ac + 1)$ multiplications. These multiplications can be parallelized for the $n_\bb$ + $n_\ac + 1$ different coefficients at a given time step, but need to be performed sequentially for the $\nsamp$ time samples due to the recurrent structure of \eqref{eq:OE_predictor_b}.

\subsection{Backward operations} 
The backward operations are illustrated in Figure \ref{fig:backprop_tf} with dashed arrows. 
In the {backward pass}, $G$ receives the vector $\adjoint{\tvec{y}} \in \mathbb{R}^{\nsamp}$ containing the partial derivatives of the loss $\loss$ w.r.t. $\tvec{y}$, namely:
\begin{equation}
\adjoint{\tvec{y}}_t = \pdiff{\loss}{\tvec{y}_t},\quad  t=0,\dots,\nsamp-1. 
\end{equation}
Given $\adjoint{\tvec{y}}$, the $G$-block has to compute the derivatives of the loss $\loss$ w.r.t. its differentiable inputs  $\bb$, $\ac$, and $\tvec{u}$. Overall, the backward operation has the following structure:

\begin{equation}
\adjoint{\bb}, \adjoint{\ac}, \adjoint{\tvec{u}} = G.{\rm backward}(\tvec{u}, \bb, \ac, \adjoint{\tvec{y}}),
\end{equation}
where 
\begin{subequations}
	\begin{align} 
	\adjoint{\bb}_j &= \pdiff{\loss}{\bb_j},\qquad j=0,\dots,n_\bb \\
	\adjoint{\ac}_j &= \pdiff{\loss}{\ac_j},\qquad j=1,\dots,n_\ac\\
	\adjoint{\tvec{u}}_\tau &= \pdiff{\loss}{\tvec{u}_\tau},\qquad \tau=0,1,\dots,\nsamp\!-\!1. \label{eq:adjsensu}
	\end{align}
\end{subequations}

\paragraph{Numerator coefficients $\bb$} 
Application of the chain rule leads to:
\begin{equation*}
\adjoint{\bb}_j = \sum_{t=0}^{\nsamp-1} \pdiff{\loss}{\tvec{y}_t} \pdiff{\tvec{y}_t}{\bb_j} = \sum_{t=0}^{\nsamp-1} \adjoint{\tvec{y}}_t \pdiff{\tvec{y}_t}{b_j}
\end{equation*}

The required sensitivities  $\sens{\bb}_j(t)  = \pdiff{\tvec{y}_t}{b_j}$, $j=0,1,\dots,n_\bb,$ can be obtained in closed-form through additional filtering operations\cite{ljung:1999system}. Specifically, an expression for $\sens{\bb}_j(t)$  is derived by differentiating 
the left and hand side of Eq. \eqref{eqn:filterA} w.r.t $\bb_j$. This yields:
\begin{align}  \label{eq:sens_b}
\A(\q) \sens{\bb}_j(t)  = u(t-j),
\end{align}
or equivalently: 
\begin{align}  \label{eq:sens_b2}
\sens{\bb}_j(t)  = \frac{1}{\A(\q)}u(t-j).
\end{align}
Thus, $\sens{\bb}_j(t)$ can be computed by filtering the input vector $u(t)$  through the linear filter $ \displaystyle \frac{1}{\A(\q)}$. Furthermore, the following condition holds: 
\begin{equation}
\label{eq:regime_b2}
\sens{\bb}_{j}(t) = \begin{cases}
\sens{\bb}_0(t-j), \;\;&t-j \geq 0\\
0,           \;\; & t-j < 0.
\end{cases}
\end{equation}
Then, one only needs to compute $ \sens{\bb}_0(t)$ by simulating  the recursive  equation   \eqref{eq:sens_b}. The other sensitivities $ \sens{\bb}_j(t)$, $j=1,\ldots,n_\bb$, are  obtained through simple shifting operations according to \eqref{eq:regime_b2}.

Exploiting expression \eqref{eq:regime_b2} for $\tilde b_j(t)$, the $j$-th component of $\adjoint{\bb}$ is obtained as:
\begin{equation}
\label{eq:backprop_b_last}
\adjoint{\bb}_j = \sum_{t=j}^{\nsamp-1} \adjoint{\tvec{y}}_t \sens{b}_0(t-j).
\end{equation}
This operation corresponds to the dot product of $\adjoint{\tvec{y}}$ with shifted version of the sensitivity $b_0(t)$.

Overall, the  computation of  $\adjoint{\bb}$ requires: ($i$) filtering $\tvec{u}$ through $\frac{1}{A(\q)}$, which entails $\nsamp n_\ac$ multiplications and ($ii$) 
the $n_{\bb}\!+\!1$ dot products defined in \eqref{eq:backprop_b_last}, totaling $\nsamp(n_{\bb}\!+\!1)- n_\bb$ multiplications. 
As for the filtering, the operations have to be performed sequentially for the different time steps due to the 
recursive structure of \eqref{eq:sens_b}. After completion of the filtering, the dot product operations may be performed in parallel.

\paragraph{Denominator coefficients $\ac$}
Following the same rationale above, we obtain a closed-form expression for the sensitivities $\sens{\ac}_j(t) = \pdiff{\tvec{y}_t}{\ac_j}$, $j=1,2,\dots,n_\ac$, 
by differentiating the left and right hand side of Eq. \eqref{eqn:filterA} with respect to $\ac_j$. This yields:
\begin{equation*}
y(t-j) + \A(\q) \pdiff{y(t)}{\ac_j} = 0,
\end{equation*}
or equivalently
\begin{equation}
\label{eq:sens_a}
\sens{\ac}_j(t) = -\frac{1}{\A(\q)}y(t-j).
\end{equation}
Then, $\sens{\ac}_j(t)$ can be obtained by filtering the output $\tvec{y}$ through the linear filter $-\frac{1}{\A(\q)}$. Furthermore, the following condition holds:
\begin{equation}
\label{eq:regime_a2}
\sens{\ac}_{j}(t) = \begin{cases}
\sens{\ac}_1(t-j+1), \;\;&t-j+1 \geq 0\\
0,           \;\; & t-j+1 < 0.
\end{cases}
\end{equation}
The $j$-th component of $\adjoint{\ac}$ is obtained as: 
\begin{equation} \label{eqn:abar}
\adjoint{\ac}_j = \sum_{t=j-1}^{\nsamp-1} \adjoint{\tvec{y}}_t \sens{a}_1(t-j+1).
\end{equation}

The back-propagation for the denominator coefficients requires: ($i$) the  filtering operation \eqref{eq:sens_a}, which involves  $\nsamp n_\ac$ multiplications; and ($ii$) 
the $n_\ac$ dot products defined in \eqref{eqn:abar}, totaling $\nsamp n_\ac - n_\ac + 1$ multiplications. 

\paragraph{Input time series  $\tvec{u}$} 
Application of the chain rule yields:
\begin{equation}
\label{eq:chainrule_u}
\adjoint{\tvec{u}}_\tau = \pdiff{\loss}{\tvec{u}_\tau} 
= \sum_{t=0}^{\nsamp-1}{\pdiff{\loss}{\tvec{y}_t} \pdiff{\tvec{y}_t}{\tvec{u}_\tau}} 
= \sum_{t=0}^{\nsamp-1}{\adjoint{\tvec{y}}_t \pdiff{\tvec{y}_t}{\tvec{u}_\tau}} 
\end{equation}
From \eqref{eq:G_conv}, the following expression for $\pdiff{\tvec{y}_t}{\tvec{u}_\tau}$ holds:
\begin{equation}
\pdiff{\tvec{y}_t}{\tvec{u}_\tau} = \begin{cases}
\tvec{g}_{t-\tau},\;\; &t-\tau \geq 0\\
0, \; & t-\tau < 0.
\end{cases}
\end{equation}
Plugging the expression above for $\pdiff{\tvec{y}_t}{\tvec{u}_\tau}$ into \eqref{eq:chainrule_u}, we obtain 
\begin{equation*}
\adjoint{\tvec{u}}_\tau = 
\sum_{t=\tau}^{\nsamp-1}\adjoint{\tvec{y}}_t 
\tvec{g}_{t-\tau} 
\end{equation*}

By definition, the expression above corresponds to the following cross-correlation operation:
\begin{equation}
\label{eq:backprop_u_corr}
\adjoint{\tvec{u}}_\tau = (\tvec{g}  \ccorr \tvec{\adjoint y})_\tau,\qquad \tau=0,1,\dots \nsamp-1.
\end{equation}
However, direct implementation of \eqref{eq:backprop_u_corr} requires a number of operations \emph{quadratic} in  $\nsamp$.

In order to obtain a more efficient solution, we observe that:
\begin{align*}
 &\adjoint{\tvec{u}}_0 = \sum_{t=0}^{\nsamp-1} \adjoint{\tvec{y}}_t \tvec{g}_t, \qquad 
  \adjoint{\tvec{u}}_1 = \sum_{t=1}^{\nsamp-1} \adjoint{\tvec{y}}_t \tvec{g}_{t-1}, \qquad  
  \adjoint{\tvec{u}}_2 = \sum_{t=2}^{\nsamp-1} \adjoint{\tvec{y}}_t \tvec{g}_{t-2}, \qquad \\
 &\dots, \qquad
 \adjoint{\tvec{u}}_{\nsamp-2} = \adjoint{\tvec{y}}_{\nsamp-2} \tvec{g}_0 + \adjoint{\tvec{u}}_{\nsamp-1}\tvec{g}_1, \qquad
  \adjoint{\tvec{u}}_{\nsamp-1} = \adjoint{\tvec{y}}_{\nsamp-1} \tvec{g}_{0}.
 \end{align*}

Since $\tvec{g}$ represents the impulse response of the filter $G(q)$, the vector $\adjoint{\tvec{u}}$ may also be obtained  by filtering the vector $\adjoint{\tvec{y}}$ in reverse time through $G(q)$, and then reversing the result, i.e., 
\begin{equation}
\label{eq:backprop_u_filt}
{\adjoint{\tvec{u}}} = \textrm{flip}\big(G(q)\textrm{flip}(\adjoint{\tvec{y}})\big).
\end{equation}

Neglecting the flipping operations, the computational cost of the backward pass for $\tvec{u}$ implemented using \eqref{eq:backprop_u_filt} is \emph{linear} in $\nsamp$. Indeed, it is equivalent to the filtering of a $\nsamp$-length vector through $G(\q)$, which requires $\nsamp(n_\bb + n_\ac + 1)$ multiplications.  

\section{Special cases of linear dynamical operators} \label{sec:specialcase}

In this section, two special cases of the linear dynamical operators, namely the  finite impulse response  and the second-order structures are analyzed. The first case is interesting as it corresponds to the convolutional block of a standard 1D CNN, which is generalized by the \Name \ architecture. The latter is useful in practice as: ($i$) higher-order systems may always be described as the sequential connection of first- and second-order dynamics; ($ii$) the coefficients of a second-order system can be readily re-parametrized in order to  enforce stability of the dynamical blocks and thus of the whole \Name\ network.   

\subsection{Finite impulse response structure}
A finite impulse response (FIR) dynamical operator has structure
\begin{equation}
G(\q) = \bb_0  + \bb_1 \q^{-1} + \dots + n_{n_\bb} \q^{-n_\bb}.
\end{equation}
In the FIR structure, there are no denominator coefficients $\ac$. Furthermore, the numerator coefficients $\bb$ correspond to the system's non-zero impulse response coefficients. 
For these reasons, the formulas derived in Section \ref{sec:layer_DL} required to define the forward and backward behavior of a general $G$-block simplify significantly in the FIR case. 
\paragraph{Forward operations}
The forward pass operation is equivalent to
\begin{equation}
\tvec{y} = G(\q) \tvec{u},
\end{equation}
which is equivalent to the convolution
\begin{subequations}
\begin{align}
\tvec{y}_i &= (\tvec{g} \conv \tvec{u})_i,  &i&=0,1,\dots,\nsamp-1\\
&= (\bb \conv \tvec{u})_i,  &i&=0,1,\dots,\nsamp-1. \label{eq:cnn_forward}
\end{align}
\end{subequations}
Using \eqref{eq:cnn_forward} for the implementation, the forward operation of a FIR $G$-block requires $\nsamp(n_\bb + 1)$ fully parallelizable multiplications. 

\paragraph{Backward operations}
The sensitivities of the block output $\tvec{y}$ with respect to the numerator coefficients $\bb$ are given by
\begin{equation*}
\sens{b}_j(t) = \pdiff{\tvec{y}_t}{b_j} = \tvec{u}_{t-j}, \qquad j=0,1,\dots,n_\bb.
\end{equation*}
Applying the chain rule, we obtain
\begin{equation*}
\adjoint{b}_j = \pdiff{\loss}{b_j} = 
\sum_{t=0}^{\nsamp} \pdiff{\loss}{\tvec{y}_t} \pdiff{\tvec{y}_t}{b_j}
= \sum_{t=j}^{\nsamp} \adjoint{\tvec{y}}_t \tvec{u}_{t-j}. 
\end{equation*}
The latter can also be written as
\begin{equation*}
\adjoint{\bb}_j =  (\tvec{u} \ccorr \adjoint{\tvec{y}})_j, \qquad j=0,1,\dots,n_\bb.
\end{equation*}
Thus, computing $\adjoint{\bb}$ requires $\nsamp  (n_\bb + 1)$ fully parallelizable multiplications.

As for the back-propagation operations with respect to the input time series ${\tvec{u}}$, Equation (21) of the main paper still holds in the FIR case. 
Applying this equation to the FIR case, we obtain:
\begin{equation*}
\adjoint{\tvec{u}}_\tau = ({\bb} \ccorr \adjoint{\tvec{y}})_{\tau}, \qquad \tau=0,1,\dots,\nsamp-1.
\end{equation*}
This operation also requires  $\nsamp  (n_\bb+1)$ fully parallelizable multiplications.

The formulas presented above for the FIR structure are very similar to the ones used in 1D-CNNs. In most deep learning frameworks, however, the cross-correlation is implemented as forward operation. 
Following the same rationale above, convolution operations appears then in the backward computations.

\paragraph{Discussion}
A significant limitation of the FIR structure is that a large number of coefficient $\bb$ is required to represent an LTI dynamics whose impulse response decays slowly.
On the other hand, all operations can be performed in parallel as they are independent for each time step, owing to the non-recurrent structure. Furthermore, the FIR representation defines by construction a stable LTI dynamics, which could have numerical advantages in training.

\subsection{Second-order structure}
A second-order dynamical operator has structure
\begin{equation}
G(\q) =  \frac{B(\q)}{\A(\q)} = \frac{b_0 + b_1\q^{-1} + b_2\q^{-1}}{1 + \ac_1\q^{-1} + \ac_2\q^{-2}}.
\end{equation}

The second-order structure is interesting as ($i$) higher-order systems may always be described as the sequential connection of first- and second-order dynamics and ($ii$) the coefficients of a second-order system can be readily re-parametrized  to  enforce stability of the block, and thus of the entire \Name\ network.   

\paragraph{System analysis}
The second-order filter $G(\q)$ above is asymptotically stable if and only if the two roots of the characteristic polynomial
\begin{equation}
P(\q) = \q^2 + \ac_1 \q + \ac_2
\end{equation}
lie within the complex unit circle. 

By applying the Jury stability criterion \cite{ogata1995discrete}, it is possible to show that this property holds in the region of the coefficient space characterized by:
\begin{subequations}
	\label{eq:jury}
	\begin{align}
	|\ac_1| &< 2\\
	|{\ac_1}| - 1 &< \ac_2 < \ac_1.
	\end{align}
\end{subequations}

\begin{figure}[h]
	\begin{center}
		
		\includegraphics[width=.4\textwidth]
		{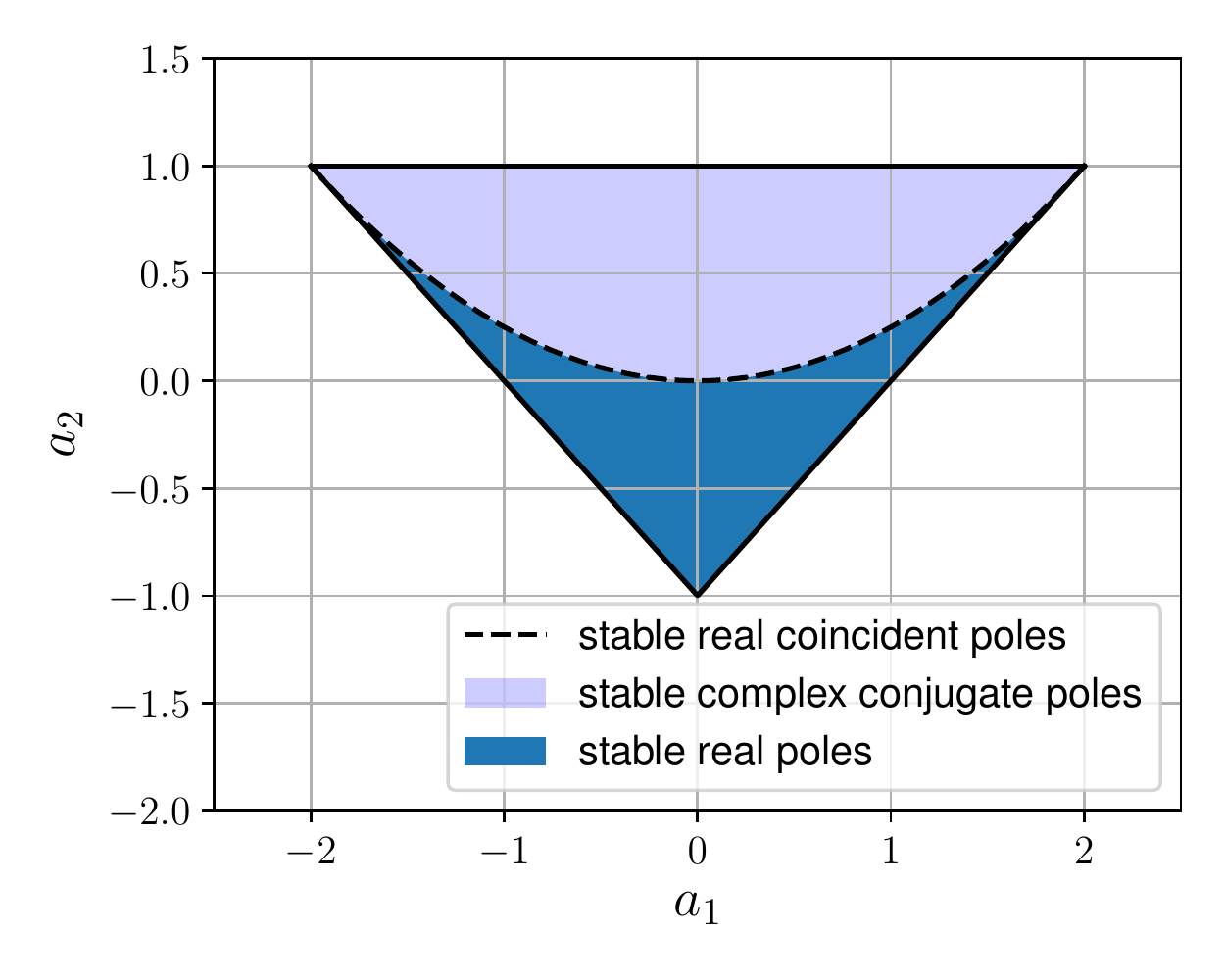}
	\end{center}
	\caption{Stability region of a second-order transfer function in the coefficient space.}
	\label{fig:stable_2ndorder}
\end{figure} 

Furthermore, the two poles are: ($i$) real and distinct for $a_2 < \frac{a_1^2}{4}$; ($ii$) complex conjugate for $a_2 > \frac{a_1^2}{4}$; and ($iii$) 
real and coincident for $a_2 = \frac{a_1^2}{4}$. The regions of interest in the coefficient space are illustrated in Figure \ref{fig:stable_2ndorder}.

\paragraph{Stable parametrization} 
An intuitive stable parametrization for second-order dynamical layers is obtained by describing the denominator $\A(\q)$ in terms of two complex conjugate (or coincident) poles:
\begin{equation*}
\A(\q) = (1 - r e^{j\beta} q^{-1}) ( 1 - r e^{-j\beta}q^{-1}) = 1 - 2r\cos\beta \q^{-1} + r^2 \q^{-2},
\end{equation*}
with magnitude $r,\;\; 0 \leq r < 1$ and phase $\beta,\;\; 0 \leq \beta < \pi$. 
Next, in order to avoid interval constraints on $r$ and $\beta$, one can further parametrize $r$ and $\beta$
in terms of unconstrained variables $\rho, \psi$ as follows: 
\begin{subequations}
	\label{eq:SOS_stable_coord}
	\begin{align*}
	r      &= \sigma(\rho)\\
	\beta  &= \pi \sigma(\psi),
	\end{align*}
\end{subequations}
where $\sigma(\cdot)$ denotes the sigmoid function and  $\rho, \psi\in \mathbb{R}$. 

The overall transformation from $\rho, \psi$ to $\ac_1, \ac_2$ is then:
\begin{subequations}
	\label{eq:reparamconj}
	\begin{align}
	\ac_1 &= - 2\sigma(\rho) \cos(\pi \sigma(\psi))\\
	\ac_2 &= \sigma(\rho)^2.
	\end{align}
\end{subequations}

Adopting this parametrization,  it is possible to train \Name\ networks that are stable by design. In practice, the trainable parameters $\rho$ and $\psi$ may be introduced in the computational graph as parents---through the differentiable transformation \eqref{eq:reparamconj}---of the coefficients $a_1$, $a_2$ of a second-order $G$-block. 
Then, the variables $\rho$, $\psi$ can be optimized (along with the numerator coefficients $b_0$, $b_1$, $b_2$ and all other 
model parameters) using standard unconstrained algorithms, with gradients computed by plain back-propagation.
The learned denominator coefficients $a_1$, $a_2$ will describe a stable second-order dynamics. 

\begin{figure}[h]
	\begin{center}
		
		\includegraphics[width=.35\textwidth]
		{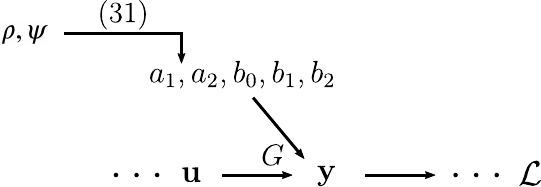}
	\end{center}
	\caption{Computational graph of a second-order $G$-block with stable re-parametrization of the denominator coefficient 
	according to transformation \eqref{eq:reparamconj}.}
	\label{fig:backprop_tf_ab_stable_mod}
\end{figure}

Figure \ref{fig:backprop_tf_ab_stable_mod} represents  the computational graph of a second-order $G$-block augmented with the 
stable re-parametrization of the denominator coefficients $a_1$, $a_2$ in terms of the unconstrained variables $\rho, \psi$.
Note that the re-parametrization formula \eqref{eq:reparamconj} included in the computational graph may be implemented 
using standard differentiable blocks readily  available in deep learning software. Thus, back-propagation 
through this operation does not entail particular difficulties.

The parametrization \eqref{eq:reparamconj} excludes however the case of two distinct real poles. 
In order to allow this system structure, a slightly more complex parametrization spanning the whole stability region \eqref{eq:jury} for the coefficients $a_1$ and $a_2$  may be used, e.g.:
\begin{subequations}
	\label{eq:reparamgen}
	\begin{align}
	\ac_1 &= 2 \tanh(\alpha_1)\\
	\ac_2 &= |2 \tanh(\alpha_1)| + (2 - |2 \tanh(\alpha_1)|)  \sigma(\alpha_2) - 1,
	\end{align}
\end{subequations}
where $\alpha_1, \alpha_2 \in \mathbb{R}$ are used as unconstrained optimization variables.

\section{Examples}
\label{sec:examples}
The effectiveness of the  \Name\ architecture is evaluated on  system identification benchmarks publicly available at the website \url{www.nonlinearbenchmark.org}.
All the codes required to reproduce the results in this section are available on the GitHub repository \url{https://github.com/forgi86/dynonet.git}

The results achieved on the Wiener-Hammerstein \cite{ljung2009wiener}, the Bouc-Wen \cite{noel2016hysteretic}, and the electro-mechanical positioning System (EMPS) \cite{janot2019data} benchmarks are presented in this paper. 
Other examples are dealt with in the provided codes.

\paragraph{Settings}
In the following examples, the \Name\ is trained by minimizing the mean square of the simulation error and  using the Adam algorithm \cite{kingma2014adam} for gradient-based optimization. The number $n$ of iterations  is chosen sufficiently large to reach a cost function plateau. The learning rate $\lambda$ is adjusted by a rough trial and error. All static non-linearities following the $G$-blocks  are modeled as feed-forward neural networks with a single hidden layer containing 20 neurons and hyperbolic tangent activation function. The numerator and denominator coefficients of the linear dynamical $G$-blocks are randomly initialized from a uniform distribution with zero mean and range $[-0.01\ 0.01]$, while the feed-forward neural network parameters are initialized according to PyTorch's default strategy. 
Note that several settings are kept constant across the benchmarks to highlight that limited tuning is needed to obtain state-of-the art identification results using the proposed \Name\ architecture.

\paragraph{Hardware setup} All computations are performed on a desktop computer equipped with an AMD Ryzen 5 1600x 6-core processor and 32 GB of RAM.

\paragraph{Metrics}
The identified models are evaluated in terms of the $\mathrm{fit}$ and Root Mean Square Error (RMSE) indexes defined as:
\begin{equation*}
\mathrm{fit} = 100\cdot \left(1- \frac{\sqrt{\sum_{t=0}^{\nsamp-1} \left(\tvec{y}^{\rm meas}_t -  {\tvec{y}}_t\right)^2} }  
{\sqrt{\sum_{t=0}^{\nsamp-1} \left(\tvec{y}^{\rm meas}_t -  {\overline{\tvec{y}}}\right)^2}}\right) (\%), 
\qquad 
\mathrm{RMSE} = \sqrt{\frac{1}{\nsamp} \sum_{t=0}^{\nsamp-1} \left(\tvec{y}^{\rm meas}_t - {\tvec{y}}_t\right)^2 },
\end{equation*}
where $\tvec{y}^{\rm meas}$ is the measured (true) output vector; ${\tvec{y}}$ is the \Name\ model's open-loop  simulated output vector;  
and $\overline{\tvec{y}}$ is the mean value of $\tvec{y^{\rm meas}}$, i.e. $\overline{\tvec{y}} = \frac{1}{\nsamp} \sum_{t=0}^{\nsamp-1} \tvec{y}^{\rm meas}_t$.

\subsection{Electronic circuit with Wiener-Hammerstein structure}
The experimental setup used in this benchmark is an electronic circuit that behaves by construction as a Wiener-Hammerstein system \cite{ljung2009wiener}. Therefore, a simple \Name\ architecture corresponding to the WH model structure is adopted. Specifically, the \Name\ model has a sequential structure defined by a SISO $G$-block with $n_a = n_b = 8$; a SISO feed-forward neural network; and a final SISO $G$-block with $n_a = n_b = 8$. 

The model is trained over $n=40000$ iterations of the Adam algorithm with learning rate $\lambda=10^{-4}$, by minimizing the  MSE on the whole training dataset ($\nsamp=100000$ samples). The total training time is 267 seconds.
On the test dataset ($\nsamp=87000$ samples), the \Name\ model's performance indexes are $\rm fit\!=\!99.5\%$ and $\rm{RMSE}\!=\!1.2$~mV. 
The measured output $\tvec{y}^{\rm meas}$ and simulated output $\tvec{y}$ on a portion of the test dataset are shown in Figure \ref{fig:result_WH}, together with the simulation error 
$\tvec{e} = \tvec{y}^{\rm meas}\! -\! \tvec{y}$.
While specialized training algorithms for WH systems may provide even superior results on this benchmark 
(the best published result \cite{schoukens2014identification} reports $\mathrm{RMSE}=0.28$~mV), the 
\Name\ model trained by plain back-propagation achieves remarkably good performance.

\subsection{Bouc-Wen system}
The Bouc-Wen is a nonlinear dynamical system describing hysteretic effects in mechanical engineering and commonly used to assess system identification algorithms. 
The example in this section is based on the synthetic Bouc-Wen benchmark\cite{noel2016hysteretic}. 
The training and test datasets of the benchmark are obtained by numerical simulation of the  differential equations:
\begin{align*}
&m_L \ddot y(t) + k_Ly(t)  + c_L \dot y(t) + z(t) = u(t)\\
&\dot z(t) = \alpha \dot y(t) - \beta\left(\gamma|\dot y(t)||z(t)|^{\nu - 1}z(t) + \delta \dot y(t)|z(t)|^\nu\right), 
\end{align*}
where $u(t)$~(N) is the input force; $z(t)$ (N) is the hysteretic force; $y(t)$ (mm) is the output displacement; and all other symbols represent fixed coefficients. The input $u(t)$ and output $y(t)$ signals are available at a sampling frequency $f_s=750$~Hz. The output $y(t)$ is corrupted by an additive band-limited Gaussian noise with bandwidth $375$~Hz and standard deviation $8 \cdot 10^{-3}$~mm. The training and test datsaset for the benchmark are generated using as input independent random phase multisine sequences containing $40960$ and $8192$ samples, respectively.

We adopt for this benchmark a \Name\ architecture with two parallel branches. The first branch has a sequential structure containing: a $G$-block with 1 input and 8 output channels; a feed-forward network with 8 input and 4 output channels; a $G$-block with 4 input and 4 output channels; and a feed-forward neural network with 4 input and 1 output channel, while the second branch consists in a single SISO $G$-block. The model output is the sum of  the two branches.
All the $G$-blocks in the first branch are third-order ($n_a\!=\!n_b\!=\!3$), while the single $G$-block in the second branch is second-order ($n_a\! =\! n_b\! =\! 2$). This model does not have a specific physical motivation and it is chosen to showcase the representational power of {\Name}. Furthermore, it does not correspond to any classic block-oriented structure previously considered in the system identification literature.

The model is trained over $n=10000$ iterations with learning rate $\lambda=2\cdot 10^{-3}$, by minimizing the MSE on the whole training dataset. On the test dataset, the model achieves a $\rm fit$ index of $93.2\%$ and a RMSE of $4.52\cdot10^{-5}$~mm.  Time traces of the measured and simulated \Name\ output on a portion of the test dataset are shown in Figure \ref{fig:result_BW}.
The results obtained by the \Name\ compare favorably with other general black-box identification methods applied to this benchmark.  
For instance,   Non-linear Finite Impulse Response (NFIR); Auto Regressive with eXogenous input (NARX); and Orthornormal Basis Function (NOBF) model structures are tested on this benchmark \cite{belz2017automatic}.
The best results are obtained with the NFIR structure ($\rm RMSE = 16.3 \,\cdot\, 10^{-5}$~mm).
Superior results are achieved  using polynomial nonlinear state-space models \cite{esfahani2017polynomial} trained with an algorithm tailored for the identification of hysteretic systems ($\rm RMSE = 1.87 \cdot 10^{-5}$~mm).

\subsection{Electro-mechanical positioning system}
The EMPS is a controlled prismatic joint, which is a common component of robots and machine tools. 
In the experimental benchmark  \cite{janot2019data}, the system input is the motor force $F$ (N) and the measured output is the prismatic  joint position $y$ (m). A physical model for the system is:
\begin{equation*}
\ddot y(t) = -\frac{F(t)}{M} - \frac{F_{d}(\dot y(t))}{M}, 
\end{equation*}
where $M$~(kg) is the joint mass and $F_{d}(t)$~(N) is the friction affecting the system (comprising both viscous and Coloumb terms). From a system identification perspective, this benchmark is challenging due to $(i)$ the unknown 
(and possibly complex) friction characteristic, $(ii)$ the marginally stable (integral) system dynamics, and $(iii)$ actuator and sensor behavior affecting the measured data. 

We adopt for this benchmark a sequential \Name\ structure comprising: a third-order $G$-block with 1 input and 20 output channels; a feed-forward neural network with 20 input and 1 output channel; and a final integrator block. 
In this architecture, the final integrator is used to model the integral system dynamics, while the other units have no 
specific physical meaning and are used as black-box model components.

\begin{figure}[h]
	\centering
	\begin{subfigure}{.33\textwidth}
		\centering
		\includegraphics[width=\linewidth]{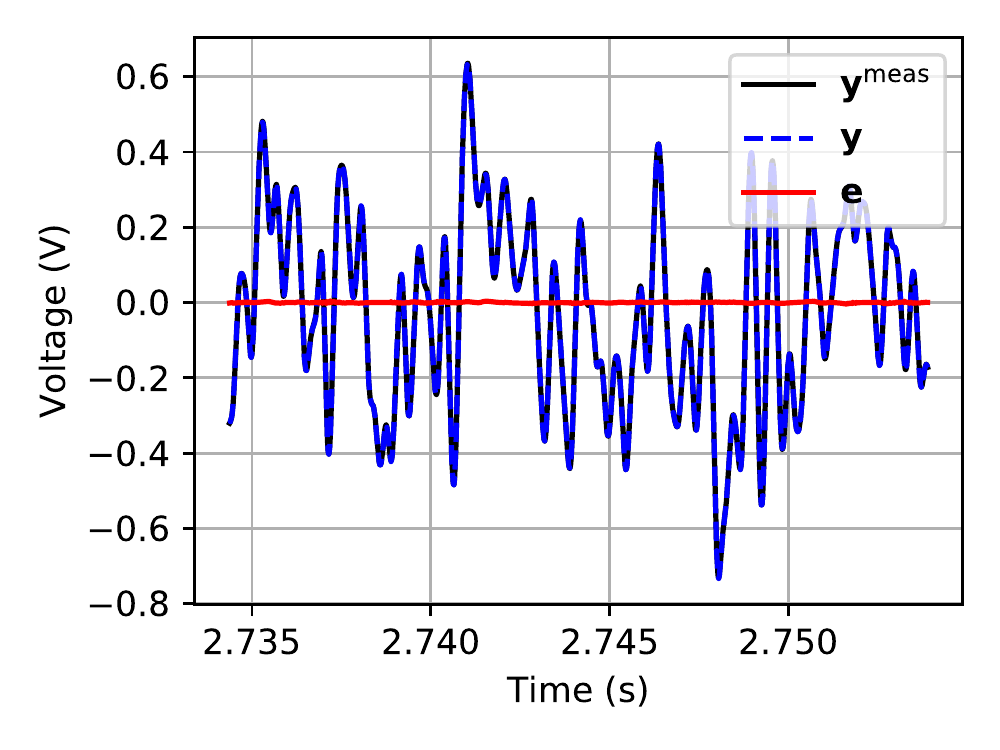}
		\subcaption{Wiener-Hammerstein circuit.}
		\label{fig:result_WH}
	\end{subfigure}%
	\begin{subfigure}{.33\textwidth}
		\centering
		\includegraphics[width=\linewidth]{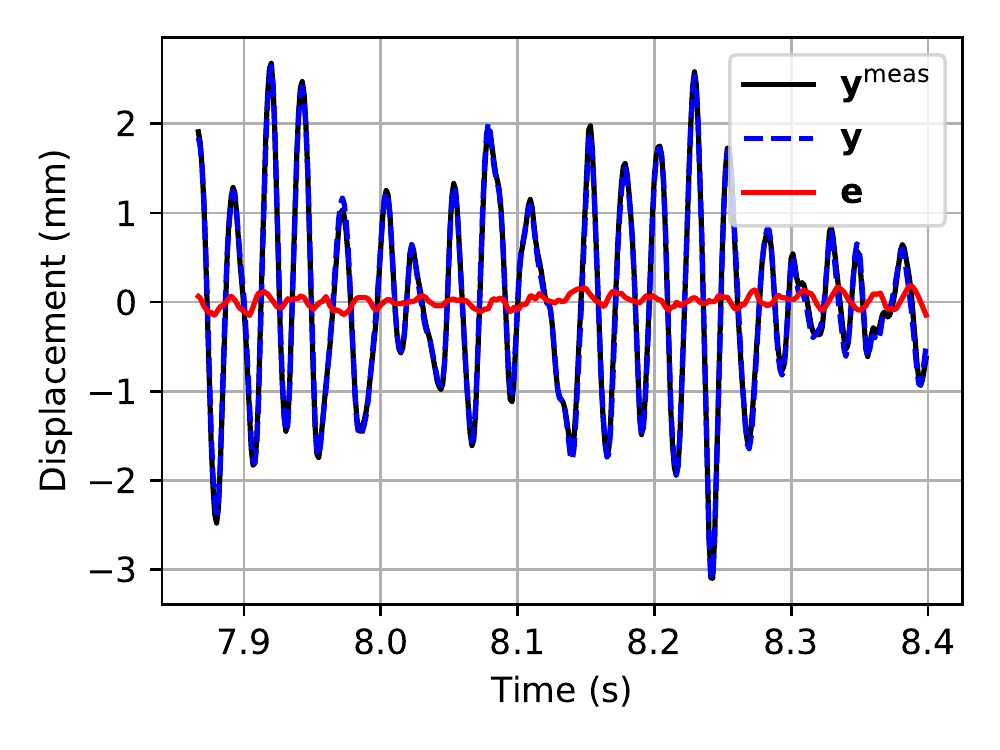}
		\subcaption{Bouc-Wen.}
		\label{fig:result_BW}
	\end{subfigure}%
	\begin{subfigure}{.33\textwidth}
		\centering
		\includegraphics[width=\linewidth]{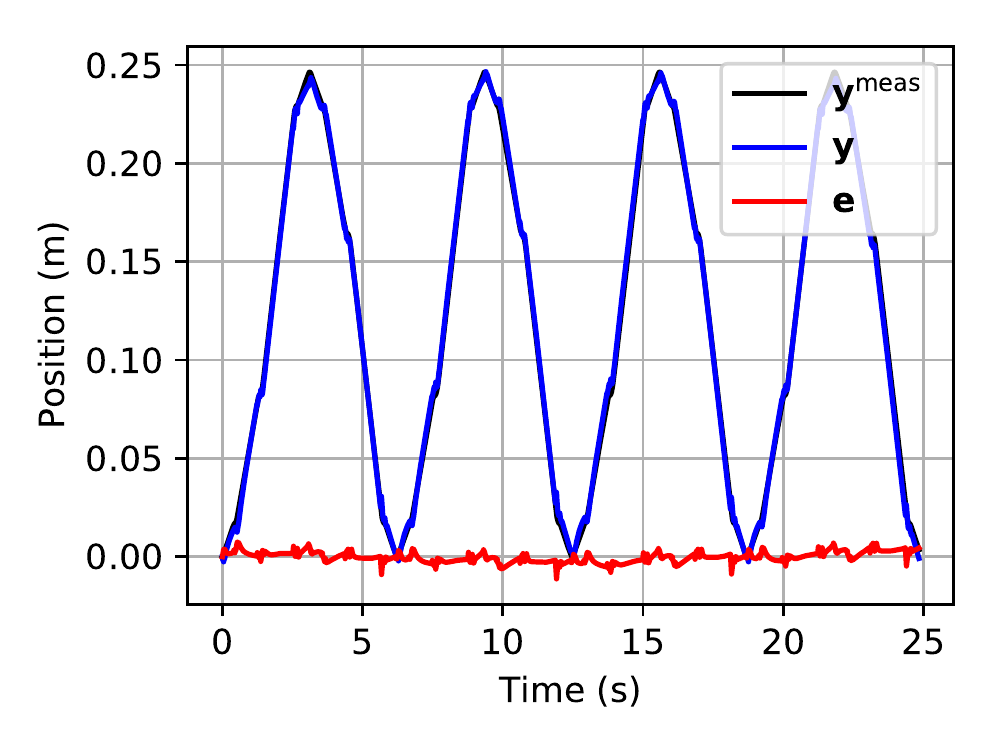}
		\subcaption{EMPS.}
		\label{fig:result_EMPS}
	\end{subfigure}%
	\caption{Measured output $\tvec{y}^{\rm meas}$ (black), \Name\ simulated output $\tvec{y}$ (blue), and  simulation error 
		$\tvec{e}=\tvec{y}-\tvec{y}^{\rm meas}$ (red)
		on the test dataset for the three benchmarks.}
	\label{fig:result}
\end{figure}

By training this \Name\ model over $n\!=\!50000$ iterations on a dataset with $\nsamp=24841$ samples with learning rate $\lambda\!=\!1\cdot10^{-4}$, we obtain on the test dataset performance indexes $\rm fit=96.8\%$ and $\rm {RMSE} = 2.64\cdot 10^{-3}$~mm. As reference, a $\rm fit=25.4\%$ for the best linear model was obtained on this benchmark \cite{janot2019data}. Time traces of the measured and simulated output on the test dataset are shown in Figure \ref{fig:result_EMPS}.

\section{Conclusions}
We have introduced \Name, a neural network architecture tailored for time series processing and dynamical system learning. The core element of  \Name \  is a linear \emph{infinite impulse response} dynamical operator described by a rational transfer function.  We have derived all the formulas required to integrate the  dynamical operator in a back-propagation-based optimization engine for end-to-end training of deep  networks. 
Furthermore,  we have analyzed the computational cost of the forward and backward operations.

The proposed case studies have shown the effectiveness and flexibility of the presented methodologies against well-known system identification benchmarks. 

Current and future research activities are devoted to the design of nonlinear  state estimators   and control strategies for systems modeled  by  \Name\ networks.

\section*{Acknowledgments}
This work was partially supported by the European H2020-CS2 project ADMITTED, Grant agreement no. GA832003.

\bibliographystyle{plain}
\bibliography{ms}

\end{document}